\documentclass[a4paper, 
10pt
]{article}

\usepackage[a4paper,left=3cm,right=3cm,top=2.5cm,bottom=2.5cm]{geometry}
\usepackage{amsmath}
\usepackage{multirow}
\usepackage[table]{xcolor}
\usepackage{authblk}
\usepackage{algorithm}
\usepackage{graphicx}

\usepackage{algpseudocode}
\usepackage{amsmath}
\usepackage{graphicx}
\usepackage{amsfonts}
\usepackage{algorithm}
\usepackage{algpseudocode}
\usepackage{enumitem}
\usepackage{cleveref}
\usepackage{booktabs}
\usepackage{calligra}
\DeclareMathAlphabet\mathzapf{T1}{pzc} {mb} {it}

\newcommand{\R}{\ensuremath{\mathbb{R}}}
\newcommand{\x}{\ensuremath{\mathbf{x}}}
\newcommand{\y}{\ensuremath{\mathbf{y}}}
\newcommand{\z}{\ensuremath{\mathbf{z}}}
\newcommand{\objdet}{\mathzapf{Obj\_Det}}
\newcommand{\basenet}{\mathzapf{basenet}}
\newcommand{\auxlayers}{\mathzapf{auxlayers}}
\newcommand{\predictor}{\mathzapf{predictor}}

\let\OLDthebibliography\thebibliography
\renewcommand\thebibliography[1]{
  \OLDthebibliography{#1}
  \setlength{\parskip}{0pt}
  \setlength{\itemsep}{0pt plus 0.3ex}
}

\title{A Proper Orthogonal Decomposition approach for parameters reduction of Single Shot Detector networks}

\author[]{Laura Meneghetti\footnote{laura.meneghetti@sissa.it}}
\author[]{Nicola Demo\footnote{nicola.demo@sissa.it}}
\author[]{Gianluigi Rozza\footnote{gianluigi.rozza@sissa.it}}

\affil{Mathematics Area, mathLab, SISSA, via Bonomea 265, I-34136
  Trieste, Italy}

\begin{document}
\maketitle
\begin{abstract}
As a major breakthrough in artificial intelligence and deep learning, Convolutional Neural Networks have achieved an impressive success in solving many problems in several fields including computer vision and image processing. Real-time performance, robustness of algorithms and fast training processes remain open problems in these contexts. In addition object recognition and detection are challenging tasks for resource-constrained embedded systems, commonly used in the industrial sector. To overcome these issues, we propose a dimensionality reduction framework based on Proper Orthogonal Decomposition, a classical model order reduction technique, in order to gain a reduction in the number of hyperparameters of the net. 
We have applied such framework to SSD300 architecture using PASCAL VOC dataset, demonstrating a reduction of the network dimension and a remarkable speedup in the fine-tuning of the network in a transfer learning context.
\end{abstract}

\section{Introduction}
\label{sec:intro}
Computer vision has attracted much research interest in recent years due to a growing attention to the problems of video analysis and image understanding. The task of estimating the location and the class of objects contained within an image is referred as \emph{object detection}~\cite{cyganek2013object, wang2016deep, zhang_objdet, liu2020deep}. In such procedure, it is not only required to classify objects contained in pictures but also to detect their position. This additional requirement to find all instances of objects leads to more complex and deep networks than traditional Convolutional Neural Networks (CNNs)~\cite{sultana_objdet, zhao_objdet, liu2020deep}, such as Faster R-CNN~\cite{faster_rcnn}, SSD~\cite{liu2016ssd}, YOLO~\cite{redmon2016you, redmon2017yolo9000}. In fact net architectures for object detection are typically composed of a base net, i.e. a CNN, and some additional structures responsible for the classification and localization of the different entities in images by drawing a labeled bounding box around each of them.\\
Training these high-dimensional deep neural networks, characterized by millions of parameters, has implications when running models in practice, especially for the possible long training time~\cite{Goodfellow, szegedy2015going} and for the required storage space. There exists object detector's implementation designed for memory constrained systems maintaining their original image classifier, but based on well-established hardware-efficient CNNs as backbone, e.g.  MobileNet~\cite{howard2017mobilenets}, ShuffleNet~\cite{Zhang2018ShuffleNetAE}, SqueezeNet~\cite{iandola2016squeezenet}, producing thus a light-weight deep neural network with reasonable performances~\cite{ssdlite, howard2017mobilenets, Mudumbi_2019, mobile_yolo, shuffle_yolo, squeezedet, shuffle_ssd}. Other approaches directly adopt the original object detection architecture employing input resizing and network pruning~\cite{han2015deep, deep_pruning, bonnaerens2021anchor}, or highly optimized base nets and convolutional feature layers~\cite{tiny_ssd} to lower network complexity.\\
Unlike what done in the aforementioned methods, to face the compute-intensive and memory-intensive issues a dimensionality reduction method is proposed, extending and adapting the one proposed for Artificial and Convolutional Neural Networks~\cite{meneghetti2021dimensionality, cui2020active}. Starting from the original structure of an object detector, the method performs a compression on the network by reducing the base net. This dimensionality reduction is carried out using Proper Orthogonal Decomposition (POD)~\cite{rozza2015book}, a well established method for model order reduction~\cite{RozzaHessStabileTezzeleBallarin2020, RozzaMalikDemoTezzeleGirfoglioStabileMola2018, SalmoiraghiBallarinCorsiMolaTezzeleRozza2016}. Such method allows for the computation of the optimal space to linearly represent the input data, here the inner features of a certain inner layer, as well as an estimation of the error introduced by projecting onto this space. We are thus retaining only a certain number of layers of the backbone and substituting the remaining part with this reduction layer, decreasing the number of hyperparameters in order to obtain benefits both in the space needed to store the architecture and in fine tuning time. It must be said that the amount of layers --- and so of information --- we discard controls the final accuracy and final dimension of the network.\\
The article is organized as follows: Section~\ref{sec:red_objdet} provides a detailed description of the reduced framework for object detectors, while in Section~\ref{sec:results} we present the results obtained by applying the proposed methodology to a benchmark deep neural network designed for the problem of object detection. Finally Section~\ref{sec:conlusion} summarizes the entire procedure and gives ideas for future developments and perspectives for our framework.


\section{Reduced Object Detector}
\label{sec:red_objdet}
\begin{figure}[htb]
\begin{minipage}[b]{1.0\linewidth}
  \centering
  \centerline{\includegraphics[width=8.5cm]{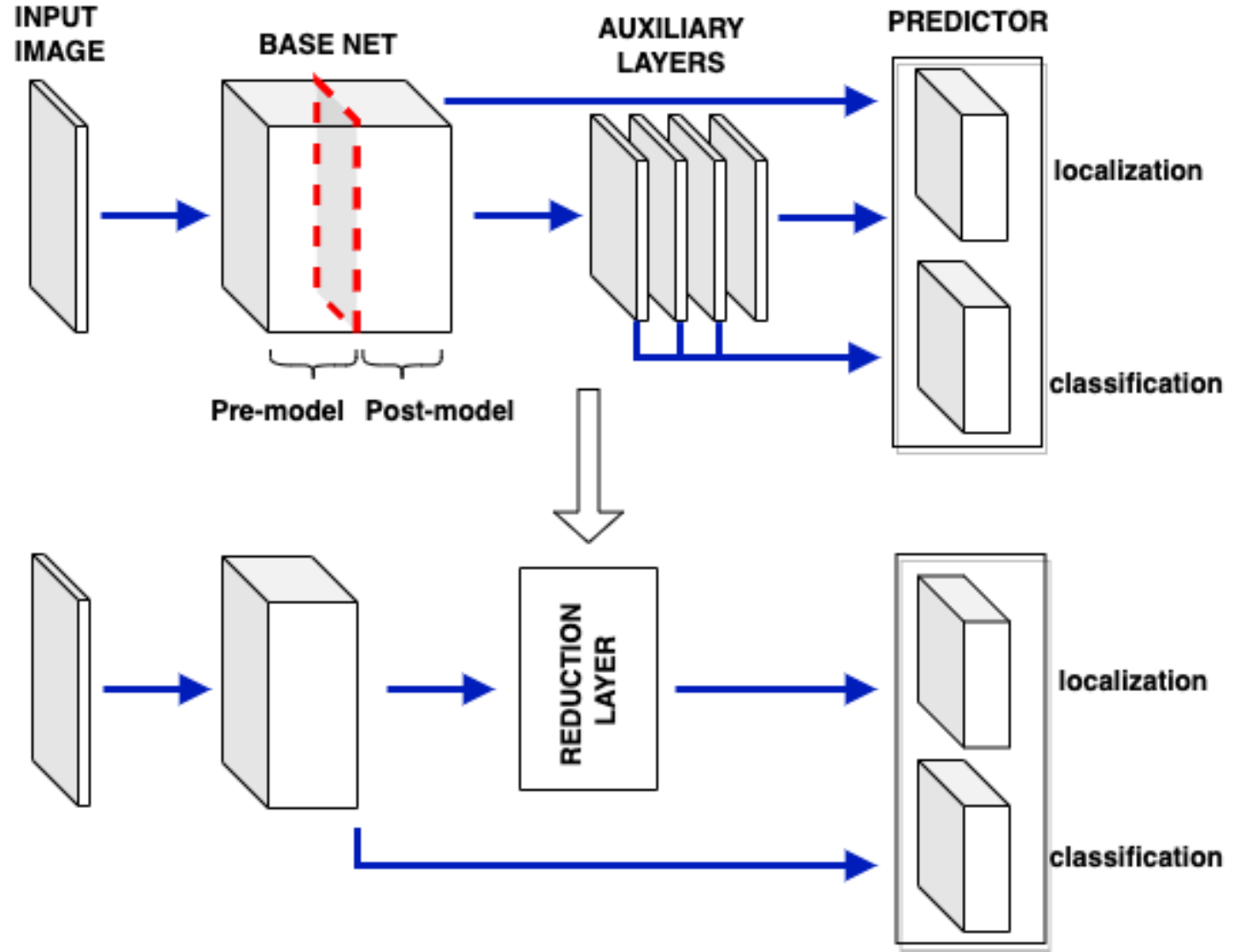}}
\end{minipage}
\caption{Graphical representation of the reduction method proposed for an object detector.}
\label{fig:red_objdet}
\end{figure}

In this section we provide the detailed description of the proposed reduced technique, that is summarized in Figure~\ref{fig:red_objdet} and in Algorithm~\ref{alg:red_objdet}. This framework is derived by the one proposed in~\cite{meneghetti2021dimensionality} to include also more complex networks, as the ones that solve the problem of object detection~\cite{sultana_objdet, zhao_objdet, liu2020deep}. The only assumption we are making is that we have a network composed by: {\it i},) a base net, a convolutional neural network extracting the low-level features; {\it ii},) some additional convolutional layers responsible for capturing the high-level features; {\it iii},) a predictor, that will output the predicted class and localization, e.g. the coordinates of the bounding box, for each object in the image.

\begin{algorithm}
\caption{Pseudo-code for the construction of the reduced object detector.}\label{alg:red_objdet}
\textbf{Inputs:} \begin{itemize}[noitemsep,topsep=0pt]
\item train dataset $\mathcal{D}_{\text{train}}=\{\x^{(0),j}, \y^j_\text{loc}, \y^j_\text{cls}\}_{j=1}^{N_{\text{train}}}$,
\item $\objdet = [\basenet, \auxlayers, \predictor]$,
\item reduced dimension $r$, 
\item index of the cut-off layer $l$,
\item a test dataset $\mathcal{D}_{\text{test}}=\{\x^i, \y^i_\text{loc}, \y^i_\text{cls}\}_{i=1}^{N_{\text{test}}}$.
\end{itemize}\vspace{4pt}
\textbf{Output:} Reduced Object Detector $\objdet^{\text{red}}$
\begin{algorithmic}[1]
\State{$\basenet_{\text{pre}}^l, \basenet_{\text{post}}^l =$ splitting\_net$(\basenet, l)$}
\State{$\x^{(l)}=\basenet^l_{\text{pre}}(\x^{(0)})$}
\State{$\z =$ reduce$(\x^{(l)}, r)$}
\State{$\hat{\y}_{\text{loc}}, \hat{\y}_{\text{cls}} =$ $\predictor(\x^{(l)}, \z)$}
\State{Training of the constructed reduced net using $\mathcal{D}_{\text{test}}$.}
\end{algorithmic}
\end{algorithm}

\subsection{Network Splitting}
At the beginning, as done in~\cite{meneghetti2021dimensionality, cui2020active}, the first part of $\objdet$, called \textit{base net}, is splitted in two different blocks: the \textit{pre-model} and the \textit{post-model}. Therefore, describing the base net as compositions of L functions $f_j: \R^{n_{j-1}} \to \R^{n_j}$, for $j=1,\dots,L$, representing the different layers of the network (e.g. convolutional with ReLU, fully connected, pooling layers), it followed that for any $1\le l < L$:
\begin{equation}
\basenet(\x^0) \equiv \basenet^l_{\text{post}}(base\_net^l_{\text{pre}}(\x^0)),
\end{equation}
where $\x^0\in\R^{n_0}$ is an input image and the pre- and post-models are defined by:
\begin{align}
\begin{split}
\basenet_{\text{pre}}^l &=  f_l \circ f_{l-1} \circ \dots \circ f_1,\\
\basenet_{\text{post}}^l &=f_L \circ f_{L-1} \circ \dots \circ f_{l+1}.
\end{split}
\end{align}
Then, the index $l$ defines the \textit{cut-off layer}, i.e. the layer at which the base net is cut. It is thus controlling how many layers of the original net, i.e. information, we are discarding.
Since its important role in the final outcome, it should be chosen carefully, usually based on considerations about the network and the dataset at hand, balancing the final accuracy and the compression ratio.

\subsection{Dimensionality Reduction}
The base net is usually followed by an additional structure responsible for the detection of the high-level features of objects, giving thus a whole understanding of the picture. In our technique, we are substituting the auxiliary layers of $\objdet$ with a linear layer performing dimensionality reduction on the pre-model output $\x^{(l)}$, as described in~\cref{alg:red_objdet}. In particular since $\x^{(l)}$ usually lies in a high dimensional space, we aim at projecting it onto a low-dimensional one by retaining only the most important directions and thus informations.\\
Based on the method presented in~\cite{meneghetti2021dimensionality}, we are exploiting the Proper Orthogonal Decomposition (POD) approach of reduce order modeling~\cite{rozza2015book} in order to decrease the number of parameters. Therefore, given  the pre-model outputs for each element of the training dataset$\{\x^{(l),i}\}_{i=1}^{N_{\text{train}}}$, we unroll them as columns of a matrix 
$\mathbf{S}=[\x^{(l),1},\dots,\x^{(l),N_{\text{train}} }]$ which we decompose
by applying the singular value decomposition, such that:
\begin{equation}\label{eq:svd_pod}
\mathbf{S} = \mathbf{\Psi}\mathbf{\Sigma}\mathbf{\Theta}^T,    
\end{equation}
where the left-singular vectors, i.e. the columns of the unitary matrix $\mathbf{\Psi}$, are the POD modes, and the diagonal matrix $\mathbf{\Sigma}$ contains the corresponding singular values in decreasing order. 
The latter indicates the contribution of all the POD modes to represent the input data, allowing for a priori selection of the number of dimension of such space.
The reduced solution $\z$ is then obtained by applying to the full solution $\x^{(l)}$ the projection matrix $\mathbf{\Psi}_r$, constructed by selecting the first $r$ modes:
\begin{equation}
\z^i= \mathbf{\Psi}^T_r\x^{(l),i},\quad \text{for}~~i=1,\dots,N_{\text{train}}.  
\end{equation}
After this reduction layer, we have then in hand the reduced representation of the pre-model output for any image in our dataset.

\subsection{Predictor}\label{sec:predictor}
To obtain the reduced network we finally need to connect the reduced features $\{\z^i\}_{i=1}^{N_\text{train}}$ to the expected outputs $\{\y^i_\text{loc}, \y^i_\text{cls}\}_{i=1}^{N_\text{train}}$. This mapping is realized by using the original classifier: several convolutional layers for localization prediction and as many for class prediction. Regarding the structural differences, we highlight that, typically, architectures for object detection use the output of several layers --- e.g. base net convolutional layers, auxiliary layers --- to predict the final output. In the reduced network, instead, the only inputs are represented by the output of the pre-model $\x^{(l)}$ and its reduced version $\z$ obtained through the reduction layer. \\
This last block of the object detector is then characterized by the presence of \textit{priors} or \textit{anchor boxes}~\cite{liu2016ssd, faster_rcnn}: fixed size reference boxes which are placed uniformly throughout the original image in order to find the correct place and sizes (width and height) of the predicted bounding boxes.
In order to discretize efficiently the space of possible predicted box length, several priors with different shapes are used for each selected input. In particular larger feature maps --- e.g. base net convolutional layers output --- have priors with smaller resolutions and are therefore ideal for detecting minor details of the image, whereas smaller feature maps --- e.g. auxiliary layers output --- will be responsible of detecting high level features, like objects' shapes. Hence, since each predictor's input is allowed to have a certain number of priors, we have performed an empirical analysis in order to find the right scale parameter, that controls the priors' resolution,  to gain comparable results with the original network. 

\section{Results}
\label{sec:results}
\begin{figure}[htb]
\begin{minipage}[b]{.48\linewidth}
  \centering
  \centerline{\includegraphics[width=4.0cm]{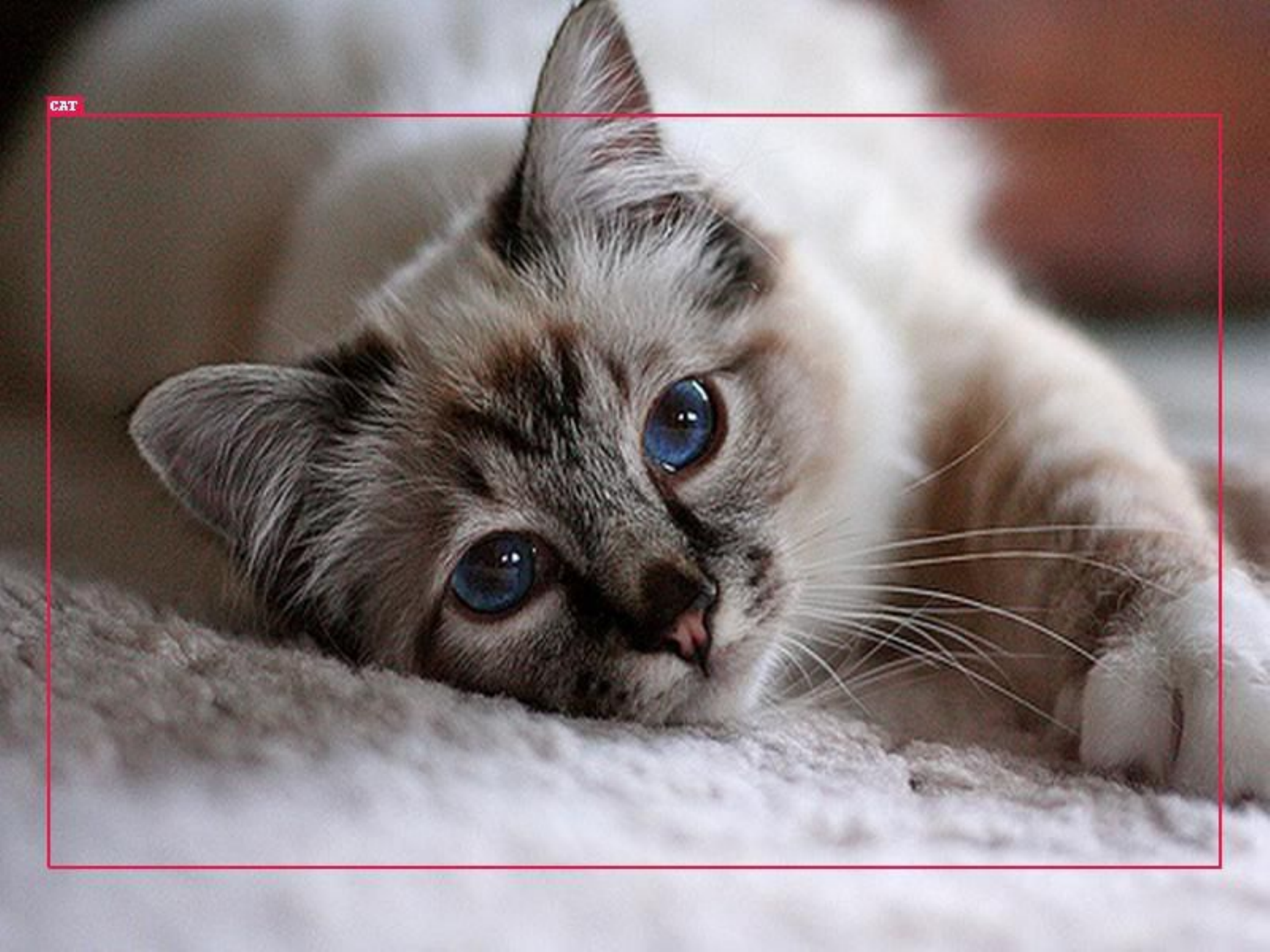}}
  \centerline{(a) SSD300}\medskip
\end{minipage}
\hfill
\begin{minipage}[b]{0.48\linewidth}
  \centering
  \centerline{\includegraphics[width=4.0cm]{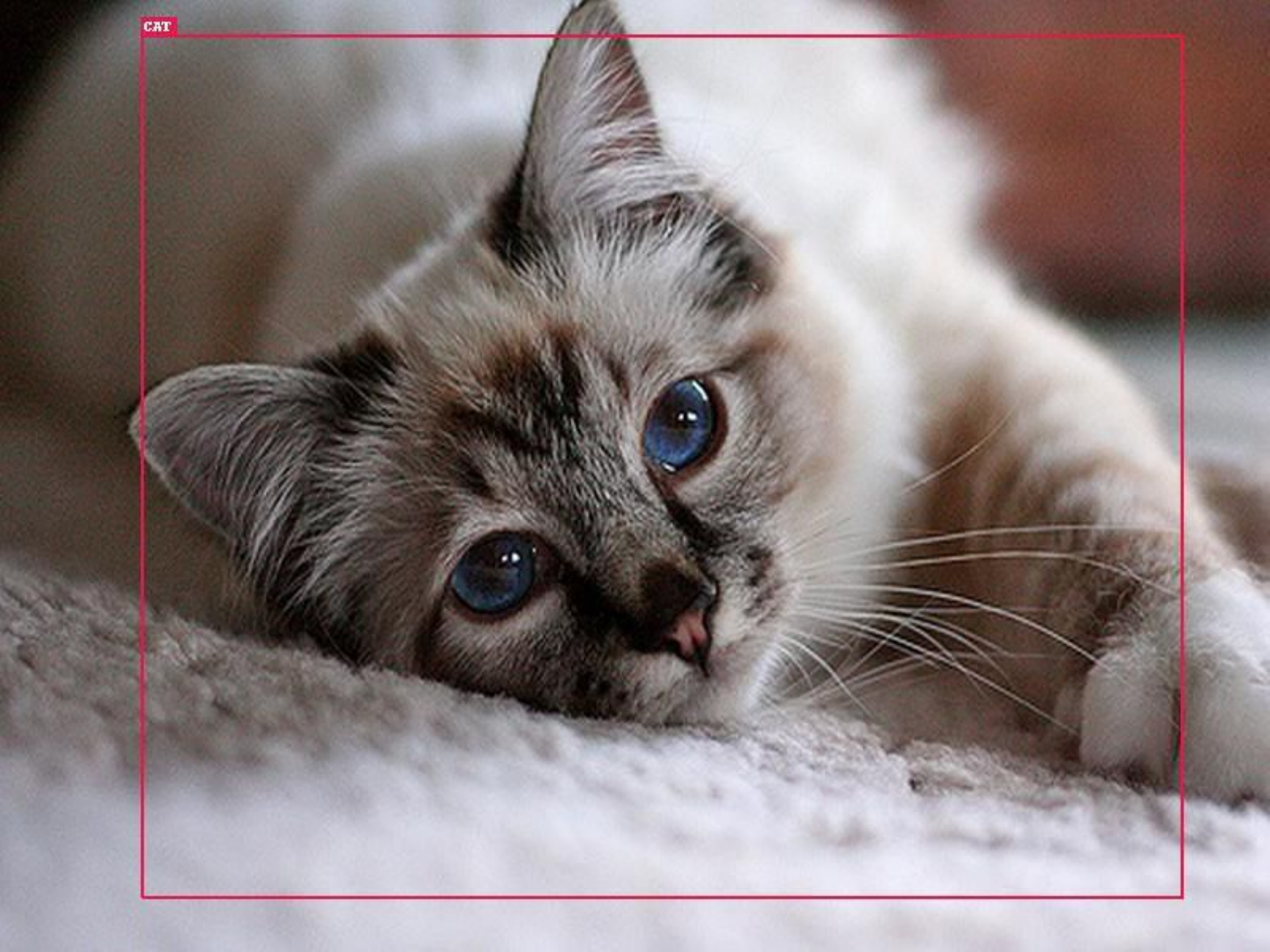}}
  \centerline{(b) Reduced SSD300}\medskip
\end{minipage}
\begin{minipage}[b]{.48\linewidth}
  \centering
  \centerline{\includegraphics[width=4.0cm]{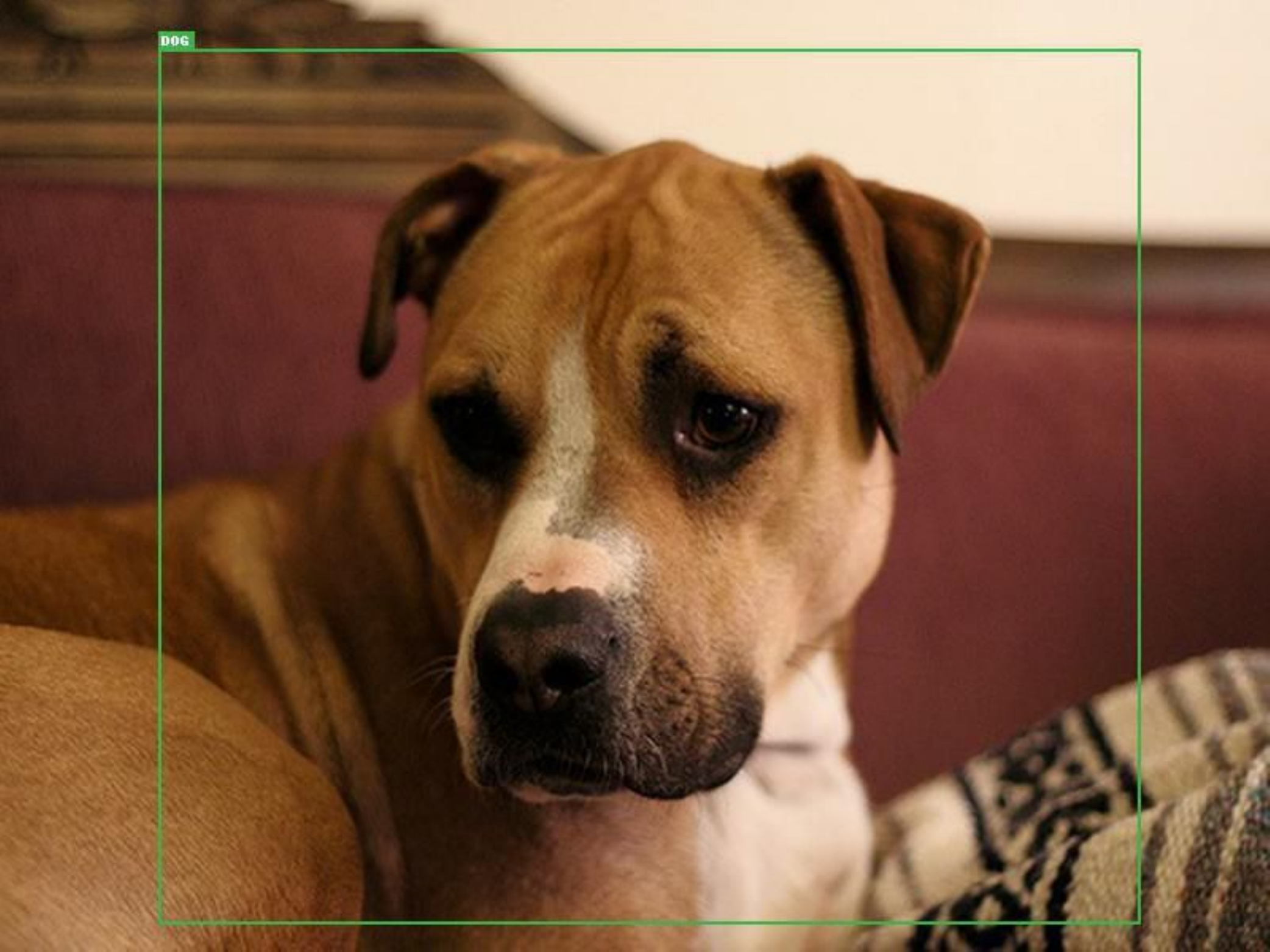}}
  \centerline{(c) SSD300}\medskip
\end{minipage}
\hfill
\begin{minipage}[b]{0.48\linewidth}
  \centering
  \centerline{\includegraphics[width=4.0cm]{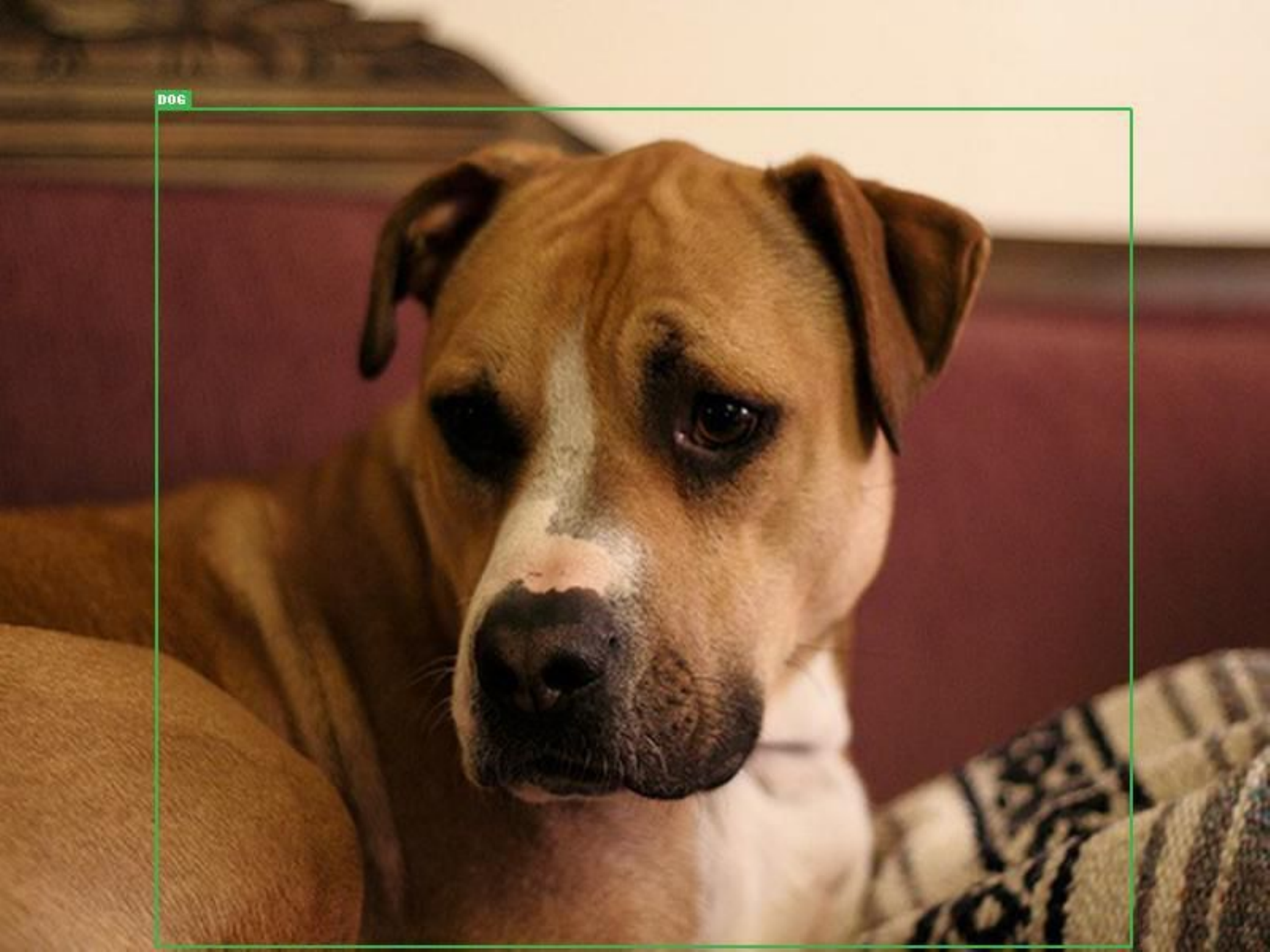}}
  \centerline{(d) Reduced SSD300}\medskip
\end{minipage}
\caption{Comparison of the results obtained using the original SSD300 and its reduced version on two test images.}
\label{fig:res_cat}
\end{figure}

We conduct experiments using SSD300~\cite{liu2016ssd} with VGG-16~\cite{vgg} as base net and the PASCAL Visual Object Classes (VOC)~\cite{Everingham10} dataset, to evaluate the ability of the full and reduced nets to detect objects in pictures. In particular, we have used both the complete PASCAL VOC and a reduced version, extracted from VOC2007, made up of 300 images subdivided in two categories: cats and dogs.
\begin{table}
\centering
\footnotesize
\caption{Results obtained with the cat-dog dataset.}\medskip
\begin{tabular}{cccc}
\toprule
\bf Network & \bf mAP & \bf Storage (Mb) &  \bf Training Time \\
\midrule
 SSD300   & 70.2\% & 91.09 & 43.5 h \\
 \midrule
 SSD300\_red & 59\% & 77.45   & 26 h\\
 \bottomrule
\end{tabular}
\label{cat_dog:res}
\end{table}
\newline
First of all, we have trained and tested the original net SSD300 with the cat-dog dataset. The results obtained are displayed in the first column of figure~\ref{fig:res_cat} and in~\cref{cat_dog:res}, where it can be seen that, after a $500$ epochs of training, the net has an overall \textit{mean Average Precision} (mAP) of $70.2\%$. In particular it reaches an accuracy of $86.6\%$ for dogs and $53.8\%$ for cats.
We have then applied the reduction method described in~\cref{sec:red_objdet} to the original architecture of SSD300, where we have not used the status of the network after training, but after initialization, i.e. with all the weights randomly initialized except those of the VGG-16, pre-trained on ImageNet~\cite{imagenet_cvpr09}. Hence, the net has been cut at layer 16, i.e. cut-off index 7, of VGG-16 and then the output of the pre-model has been projected onto a reduced space of dimension $r=50$, in analogy with~\cite{meneghetti2021dimensionality, cui2020active}. As highlighted in~\cref{sec:predictor} the inputs to the predictor part are different from the full model. In the case of SSD300 these correspond to: the output of the third layer in the fourth convolutional block conv4\_3, the final output of the net conv7, the outputs of each block of the auxiliary layers conv8\_2, conv9\_2, conv10\_2, conv11\_2. In our reduced network the only inputs used for predictions are the output of the pre-model, i.e. the output of layer conv3\_3, and its reduced version, the output of the reduction layer. In this way, the number of priors we are taking into account is less than before: 5782 instead of 8732 for the full net. We have then adapted the scaling factor to our case, choosing two different scales: 0.1 and 0.9, i.e. one that takes into account smaller objects (10\% of the picture) and one for bigger objects (90\% of the picture). Despite these changes connected with inputs and default boxes, the predictor block remains the same as the original SSD300.\\
The reduced net has firstly been trained with the cat-dog dataset for $500$ epochs. Table~\ref{cat_dog:res} summarizes the results obtained with our reduced method. As can be seen, in this case the mAP is decreased by 11\% with respect to the original net: the accuracy for the categories dogs and cats are now $67.7\%$ and $50.3\%$ respectively. Figure~\ref{fig:res_cat} shows then how the results obtained with the full and the reduced net are comparable.
\begin{table}
\centering
\footnotesize
\caption{Results obtained with PASCAL VOC.}\medskip
\begin{tabular}{cccc}
\toprule
\bf Network & \bf mAP & \bf Storage (Mb) &  \bf Training Time \\
\midrule
 SSD300   & 77.8\% & 100.23 & 48 h \\
 \midrule
 SSD300\_red & 39\% & 76.23   & 18 h\\
 \bottomrule
\end{tabular}
\label{pascal_voc:res}
\end{table}
\newline
We have then performed training using all the 20 categories of PASCAL VOC. We have trained the original SSD300 and our reduced version for 232 epochs\footnotemark\footnotetext{We have reduced the number of epochs for computational limits.}. As described in~\Cref{pascal_voc:res}, in this case the lightweight version is not as accurate as the full network. In fact, there is an overall reduction in the mAP of $38\%$.\\
Despite this decreased precision, we have achieved a reduction of memory storage, $15\%$ in the first case and $22\%$ in the second one, and a halving of the training time. We highlight that such compression is not remarkable as the one shown in~\cite{meneghetti2021dimensionality} since in this case the larger dimension of the pre-model output have led a huge matrix for storing the POD modes, limiting a lot the potential space gain. However, beside the space reduction, our proposed approach has allowed to accelerate the fine-tuning (performed also in the full network to optimize the auxiliary and classifier layers) of the network in the reduced version: assuming the weights of base net layers are already pre-trained\footnote{Mandatory for building the POD space.}, the minor number of hyperparameters results in a faster optimization task. 
Reduction of CNN becomes then not only a technique to compress the architecture dimension, but also accelerating the learning step of the studied network at the cost of marginally decreased accuracy.

\section{Conclusion}
\label{sec:conlusion}
We have applied an already explored methodology for neural network compression to one of the most common architecture for object detection, the SSD300 model. Without loosing of generality, the approach consists of a (linear) reduction layer derived from POD used to connect the selected layers of the base net (pre-model) with the original predictor block. The comparison between the original network and its reduced version shows a faster training and less required space for the latter, but at the cost of a diminished accuracy.
The obtained results shows an intuitive trend between the reduction and the accuracy with respect to the complexity of the problem, in this case roughly quantified by the number of image classes. Keeping indeed the same reduced architecture for the two different datasets ($2$ and $20$ classes), we have seen a drastic difference in the average accuracy, that demonstrates the ability of the proposed method to automatically detect redundant and superfluous information.
Future works will better explore the trade-off between complexity of the problem and compression of the network, aiming hopefully to detect an automatic way to compute the cutoff layer depending on the problem in hand.

Regarding the impact of the reduced network in memory instead, this application has demonstrated that, dealing with high-dimensional features in the pre-model, can vanishing the compression of the network. The POD modes, which dimension is equal to pre-model features, require a huge amount of space to be stored, potentially reaching the full network as in this case. One possible solution to overcome this limit is the employment of hyperreduction technique or POD variants to minimize the quantity of data that has to be saved.

\bibliographystyle{abbrv}
\bibliography{biblio}

\end{document}